\documentclass[runningheads]{llncs}

\usepackage{eccv}

\usepackage{eccvabbrv}

\usepackage{graphicx}
\usepackage{booktabs}

\usepackage[accsupp]{axessibility}  % Improves PDF readability for those with disabilities.

\usepackage{hyperref}

\usepackage{orcidlink}

\begin{document}
% \renewcommand\thelinenumber{\color[rgb]{0.2,0.5,0.8}\normalfont\sffamily\scriptsize\arabic{linenumber}\color[rgb]{0,0,0}}
% \renewcommand\makeLineNumber {\hss\thelinenumber\ \hspace{6mm} \rlap{\hskip\textwidth\ \hspace{6.5mm}\thelinenumber}}
% \linenumbers
%\pagestyle{headings}
%\mainmatter
%\def\ECCV16SubNumber{13663}  % Insert your submission number here

%\title{Atten-DualFace: Enhancing Face Recognition via Semantic-Attention and Inter-class Repulsive Margins} 

%\title{Learning to Attract and Repel: Dual Quality Margin Learning for Face Recognition (DQM-Face)}

%\titlerunning{ECCV-26 submission ID %\ECCV16SubNumber}
%\authorrunning{ECCV-26 submission ID %\ECCV16SubNumber}

%\author{Anonymous ECCV submission}
%\institute{Paper ID \ECCV16SubNumber}

% ---------------------------------------------------------------
% TODO REVIEW: Replace with your title
\title{Learning to Attract and Repel: Dual Quality Margin Learning for Face Recognition (DQM-Face)} 

% TODO REVIEW: If the paper title is too long for the running head, you can set
% an abbreviated paper title here. If not, comment out.
\titlerunning{DQM-Face: Dual Quality Margin Learning for Face Recognition}

% TODO FINAL: Replace with your author list. 
% Include the authors' OCRID for the camera-ready version, if at all possible.

\author{El Ouanas Belabbaci\inst{1}\orcidlink{0009-0001-1119-7668} \and
Bhavesh Wani\inst{2,}\orcidlink{0009-0003-6185-4886} \and
Philipp Terhörst\inst{2}\orcidlink{0000-0001-8250-5712}}

%\author{El Ouanas Belabbaci\inst{1}\orcidlink{0009-0001-1119-7668} \and
%Wani Bhavesh Gajendra\inst{2,3}\orcidlink{0009-0003-6185-4886} \and
%Terhörst Philipp\inst{3}\orcidlink{0000-0001-8250-5712}}

\institute{Laboratoire d’Ingénierie des Systèmes Intelligents et Communicants (LISIC), Faculty of Electrical Engineering, University of Sciences and Technology Houari Boumediene (USTHB), Algiers, Algeria \and
Johannes Gutenberg University Mainz, Germany}

% TODO FINAL: Replace with an abbreviated list of authors.
\authorrunning{Belabbaci et al.}
% First names are abbreviated in the running head.
% If there are more than two authors, 'et al.' is used.

% TODO FINAL: Replace with your institution list.
%\institute{Princeton University, Princeton NJ 08544, USA \and
%Springer Heidelberg, Tiergartenstr.~17, 69121 Heidelberg, Germany
%\email{lncs@springer.com}\\
%\url{http://www.springer.com/gp/computer-science/lncs} \and
%ABC Institute, Rupert-Karls-University Heidelberg, Heidelberg, Germany\\
%\email{\{abc,lncs\}@uni-heidelberg.de}}

\maketitle

%%%%%%%%%%%%%%%%

\begin{abstract}

Face recognition in unconstrained environments remains high\-ly challenging due to diverse and extreme variations encountered in real-world scenarios. To mitigate these effects, existing margin-based approaches model sample quality through feature magnitude. However, magnitude-based modeling alone is susceptible to identity-agnostic noise, which can degrade the reliability and discriminative power of learned representations.
In this paper, we propose Dual Quality Margin Learning for Face Recognition (DQM-Face), a novel framework that enables refined attraction and repulsion dynamics during representation learning. Our approach unifies conventional magnitude-based quality estimation with a newly introduced semantic quality learning mechanism, realized via squeeze-and-excitation semantic attention. By jointly leveraging magnitude and semantic cues, we construct enhanced quality-aware margins that adaptively strengthen intra-class compactness through improved attraction during learning.
To further enhance inter-class discrimination, we introduce a repulsion margin formulation that explicitly enlarges inter-class separation. The unified integration of semantic quality modeling with dual attraction–repulsion margin optimization results in a more structured and discriminative feature geometry.
Extensive experiments on multiple challenging benchmarks demonstrate that DQM-Face consistently outperforms state-of-the-art face recognition methods. Moreover, we show that the quality learned for margin optimization is highly effective for face image quality assessment within the proposed framework, demonstrating that the learned quality signal is intrinsically aligned with the recognition objective. The code is publicly available: \url{https://github.com/RAIB-group/DQM-Face}


\end{abstract}

\section{Introduction}

Face recognition (FR) remains one of the most widely deployed biometric technologies due to the collectability, universality, and high discriminative potential of the facial trait~\cite{ijcb26-2}. With the proliferation of applications in security, surveillance, and authentication, these systems are increasingly used in unconstrained real-world scenarios~\cite{DBLP:journals/ijon/WangD21a}. In such environments, recognition robustness is frequently challenged by extreme variations, such as in pose, illumination, occlusion, expression, or age, often resulting in degraded representations and a higher likelihood of false matches.

To analyze and mitigate the effects of these variations, face image quality assessment (FIQA) has emerged as a key research direction. FIQA methods aim to estimate the utility of a face image for recognition, enabling systems to quantify image reliability and improve matching performance by rejecting low-quality samples. Parallel research in representation learning has focused on developing margin-based loss functions~\cite{deng2019arcface,wang2018cosface,liu2017sphereface}, which improve feature discriminability by enforcing intra-class compactness and inter-class separation. More recently, approaches that integrate quality information directly into training objectives~\cite{meng2021magface,kim2022adaface,terhorst2023qmagface} have shown promising advances, allowing networks to adapt decision making and learning behavior based on sample difficulty.
However, current quality-aware margin formulations predominantly rely on feature magnitude as a proxy for image quality. While effective to some extent, magnitude-based estimation remains identity-agnostic: factors such as blur, occlusion, or illumination can artificially inflate embedding norms, leading to inaccurate quality predictions. Furthermore, existing methods primarily focus on intra-class compactness without explicitly enforcing inter-class repulsion, leaving a residual risk of feature overlap across identities in challenging conditions~\cite{song2024coreface,liu2022sphereface}.

To address these limitations, we propose \textbf{Dual Quality Margin Learning for Face Recognition (DQM-Face)}, a unified framework that incorporates dual quality modeling and dual margin optimization. Our method combines conventional magnitude-based quality estimation with a novel semantic quality branch implemented via squeeze-and-excitation attention. This dual formulation enables identity-aware, context-dependent quality assessment that better reflects the true recognizability of each sample. Based on these quality cues, DQM-Face introduces a dual-margin optimization mechanism that adaptively scales the target (attractive) margin according to sample quality while simultaneously applying an explicit inter-class repulsion margin. This synergy ensures compact clustering of high-quality samples while maintaining sufficient angular separation between different identities.

Extensive experiments on both image-based and video-based benchmarks demonstrate that DQM-Face consistently achieves state-of-the-art performance. The proposed approach delivers robust recognition under severe variations in pose, illumination, and image quality, ranking first or second across all major protocols, including LFW, CFP-FP, AgeDB-30, CPLFW, IJB-B, and IJB-C.

\noindent In summary, the main contributions of this work are as follows:  
(1) We propose a unified Dual Quality Margin Learning (DQM-Face) framework that integrates sample quality modeling and margin-based optimization for more robust face representation learning.  
(2) We introduce a dual quality modeling mechanism that fuses magnitude-based and semantic attention–based quality cues, producing more reliable and identity-sensitive quality estimates. These prove to be highly effective for face image quality assessment within the proposed framework and face recognition in general.
(3) We design a dual-margin optimization strategy combining an adaptive attraction margin and a curriculum-based repulsion margin to enhance both intra-class compactness and inter-class separability.  
(4) We achieve consistent state-of-the-art results across multiple challenging benchmarks, validating the effectiveness and generalization ability of the proposed approach.

\section{Related Work}

\subsection{Face Image Quality}
Face Image Quality Assessment (FIQA) aims to estimate the utility of a face image for recognition, thereby predicting the expected impact on recognition performance. Early FIQA methods were based on predefined image quality metrics, whereas recent methods learn quality predictors aligned with face recognition performance. Uncertainty-driven methods, including SER-FIQ \cite{Terhorst2020SERFIQUE}, estimate quality by measuring embedding stability under stochastic perturbations. Recognition-aware learning-based methods, including CR-FIQA \cite{boutros2023cr}, CLIB-FIQA \cite{Ou_2024_CVPR}, GraFIQs \cite{kolf2024grafiqs}, FROQ \cite{babnik2025froq}, and FaceQAN \cite{babnik2022faceqan}, learn quality predictors directly from face recognition supervision, such as genuine–impostor score separability or recognition score distributions, in supervised or semi-supervised settings. Diffusion-based approaches, such as DiffIQA \cite{babnik2023diffiqa} and eDiffIQA \cite{10468647}, use diffusion models to estimate image utility based on reconstruction consistency or denoising stability. While these methods provide valuable indicators of facial image quality, they are typically used as external assessors and not integrated into the feature learning process itself.

\subsection{Face Recognition}
The introduction of margin-based softmax losses such as SphereFace~\cite{liu2017sphereface}, CosFace~\cite{wang2018cosface}, and ArcFace~\cite{deng2019arcface} established the foundation for modern FR by enforcing margins in the normalized embedding space. These formulations improve intra-class compactness and inter-class separability, making them standard practice for identity-discriminative feature learning. However, fixed-margin formulations treat all samples equally, ignoring variations in sample difficulty and image quality. 
To address these limitations, several methods have explored adaptive margin mechanisms and optimization curricula. CurricularFace~\cite{huang2020curricularface} applied curriculum learning to progressively increase margin difficulty during training, while ElasticFace~\cite{boutros2022elasticface} introduced stochastic margin sampling from a Gaussian distribution to improve regularization and model generalization. GroupFace~\cite{kim2020groupface} extended margin-based learning by incorporating group-aware supervision to capture structured intra-class relations across identities. More recently, CoReFace~\cite{song2024coreface} employed contrastive regularization to explicitly enhance inter-class separation and stabilize hard-sample optimization. QCFace~\cite{doan2025qcface} further revisited quality-driven representation learning by integrating content consistency constraints within the margin optimization process.
Despite these advances, most methods remain agnostic to explicit image quality characteristics, leaving robustness under unconstrained conditions an open challenge.

\subsection{Quality-Aware Face Recognition}

Integrating face image quality directly into the feature learning and decision-making process has proven to be an effective strategy for achieving robust performance in unconstrained environments. MagFace~\cite{meng2021magface} introduced a unified learning framework in which the embedding magnitude serves as an implicit indicator of image quality, enabling adaptive margin scaling during training and resulting in more discriminative representations. QMagFace~\cite{terhorst2023qmagface} extended this concept by incorporating the magnitude-derived quality information into the decision-making process, improving robustness under challenging conditions. AdaFace~\cite{kim2022adaface} further refined MagFace’s quality–embedding principle by learning the relationship between feature magnitude and verification performance through adaptive quality weighting. Although these methods achieve strong performance, their magnitude-only quality proxies remain sensitive to identity-agnostic noise such as blur, occlusion, or illumination variations, which can distort the underlying quality estimation.

In contrast, our proposed DQM-Face integrates both magnitude-based and semantic quality modeling within the learning objective itself. By combining feature magnitude with semantic attention–based cues and jointly optimizing attraction and repulsion margins, DQM-Face produces identity-aware, quality-adaptive embeddings that maintain high discriminability even under severe data variations.

\section{Methodology}
In this section, we present the proposed Dual Quality Margin Learning framework (DQM-Face) designed to jointly account for sample quality and class separability in face representation learning. Building upon the standard margin-based softmax formulation, our method extends traditional magnitude-aware approaches by integrating dual quality modeling and dual margin optimization. Specifically, we combine magnitude- and semantic-based quality estimation to derive an adaptive attraction margin that promotes intra-class compactness, while introducing an explicit repulsion margin that enhances inter-class discriminability. The resulting Dual Quality Margin (DQM) loss unifies these components within a single optimization objective, enabling the model to learn structured and quality-aware embeddings suitable for robust recognition under unconstrained conditions.

\begin{figure}
    \centering
    \includegraphics[width=0.99\linewidth]{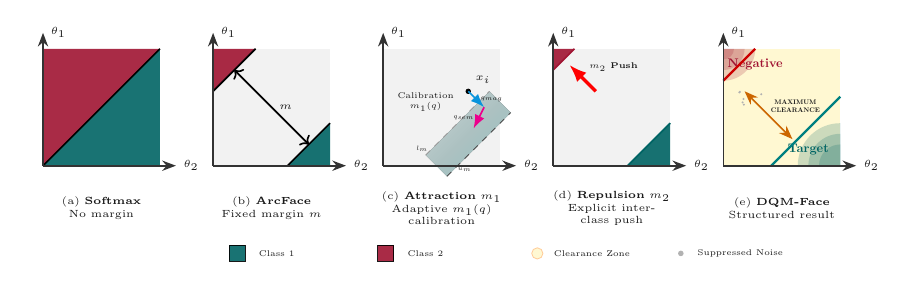}
    \caption{\textbf{Geometrical interpretation -} Feature space and decision boundaries are shown for (a) Softmax baseline with inter-class overlap, (b) ArcFace with a fixed margin $m$, (c) our adaptive attraction margin $m_1$ calibrated by dual-quality fusion, (d) explicit inter-class repulsion $m_2$ pushing non-target features away, and (e) the resulting structured feature space characterized by maximized intra-class compactness and a wide Clearance Zone.}
    \label{fig:concept}
\end{figure}

Figure \ref{fig:concept} shows an overview of the effect of different loss functions on decision boundaries in the feature space. While the standard softmax leads to an overlap zone (a) and ArcFace works with an unflexible and fixed margin (b), DQM-Face leads to a highly structured feature space with a wide clearance zone and inter-class compactness (e). This is achieved through the adaptive attraction (c) and the explicit repulsion margin (d).

\subsection{Dual Quality Margin Loss}

In face recognition, the most commonly used loss formulation is the margin-based softmax loss \cite{deng2019arcface}
\begin{equation}
\mathcal{L}_{base} = -\frac{1}{N}\sum_{i=1}^{N} \log \frac{e^{s \cdot \phi(\theta_{y_i}, m)}}{e^{s \cdot \phi(\theta_{y_i}, m)} + \sum_{j \neq y_i} e^{s \cos \theta_j}},
\label{eq:base_loss}
\end{equation}
where $\mathbf{x}_i \in \mathbb{R}^d$ denotes the face representation (template) of the $i$-th sample belonging to class $y_i$, $\theta_j = \arccos(W_j^T \mathbf{x}_i / s)$ describes the angle between the representation $\mathbf{x}_i$ and the class center $W_j \in \mathbb{R}^d$ for class $j$ and $\phi(\theta_{y_i}, m)$ describes a margin function. 
For $\phi(\theta_{y_i}, m) = \cos(\theta_{y_i} + m)$, this leads to the ArcFace loss \cite{deng2019arcface} and for $\phi(\theta_{y_i}, m) = \cos \theta_{y_i} - m$, this leads to the CosFace loss \cite{wang2018cosface}.
These formulations enforce intra-class compactness by increasing the margin between samples and their ground-truth centers, but they lack mechanisms to account for sample quality or explicitly separate inter-class features.

In this work, we propose a Quality Magnitude Combination (QMC) loss
\begin{align}
    \mathcal{L}_{QMC} =\mathcal{L}_{DQM}(x_i, m_1(q_{mag}, q_{sem}), m_2 + \lambda_g \mathcal{L}_{mag}(\|\mathbf{x}_i\|),
\end{align}
consisting of our novel Dual Quality Margin (DQM) loss and MagFace \cite{meng2021magface} loss.
The MagFace loss $\mathcal{L}_{mag}(\|\mathbf{x}\|) = \frac{1}{\|\mathbf{x}\|} + \frac{\|\mathbf{x}\|}{u_a^2},
\label{eq:mag_reg}
$  is a regularization term that enforces magnitude
monotonicity and that the length of the representation encodes its quality.
The Dual Quality Margin (DQM) loss
\begin{align}
    \mathcal{L}_{DQM} = -\frac{1}{N} \sum_{i=1}^{N} \log \frac{e^{s \cdot \cos(\theta_{y_i} + m_1(q_{mag,i}, q_{sem,i}))}}{e^{s \cdot \cos(\theta_{y_i} + m_1(q_{mag,i}, q_{sem,i}))} + \sum_{j \neq y_i} e^{s \cdot \cos(\theta_j - m_2)}} \label{eq:DQM}
\end{align}
consists of the attractive quality margin $m_1(q_{mag,i}, q_{sem,i})$ based on magnitude and semantic qualities, as well as the explicit repulsion margin $m_2$.
The gradient of the combined loss $\mathcal{L}_{QMC}$ demonstrates its effect well during training
\begin{equation*}
\frac{\partial\mathcal{L}_{DQM}}{\partial\mathbf{x}_i} = 
\underbrace{\frac{\partial\mathcal{L}_{DQM}}{\partial\cos\theta_{y_i}} \cdot \frac{\partial\cos\theta_{y_i}}{\partial\mathbf{x}_i}}_{\text{intra-class pull}} + 
\underbrace{\sum_{j\neq y_i} \frac{\partial\mathcal{L}_{DQM}}{\partial\cos\theta_j} \cdot \frac{\partial\cos\theta_j}{\partial\mathbf{x}_i}}_{\text{inter-class push}} + 
\underbrace{\lambda_g \frac{\partial\mathcal{L}_{mag}}{\partial\|\mathbf{x}_i\|} \cdot \frac{\partial\|\mathbf{x}_i\|}{\partial\mathbf{x}_i}}_{\text{quality regularization}}
\label{eq:gradient}
\end{equation*}
The first term pulls the representation toward its ground-truth center with strength modulated by the adaptive margin $m_1$. The second term pushes it away from all non-target centers with a repulsive force determined by $m_2$. The third term enforces the quality-magnitude correlation, ensuring that the feature norm accurately reflects the recognizability of the sample.
In the following, we describe the contributions of both margins in more detail.

\subsection{Attraction Margin Learning with Quality Awareness}

\subsubsection{Magnitude-based Quality Learning}

%For the magnitude-based quality learning, following \cite{meng2021magface}, the feature norm given by the operational bounds is 
%\begin{equation}
%q_{mag} = \frac{\min(\max(r, l_a), u_a) - l_a}{u_a - l_a},
%\label{eq:mag_score}
%\end{equation}
%where $l_a$ and $u_a$ are the lower and upper bounds of the feature norm, respectively. The operation $\min(\max(r, l_a), u_a)$ ensures that the feature norm is projected into the interval $[l_a, u_a]$, preventing extreme values from dominating the quality estimation. This linear normalization maps the bounded feature norm to a quality score. However, this does not cover identity-agnostic noise, which can degrade the discriminativeness of the representations.

For the magnitude-based quality learning, following \cite{meng2021magface}, the feature norm \(\|\mathbf{x}_i\|\) is used as the magnitude-based quality indicator. Given the operational bounds, the quality score is defined as 
\begin{equation}
q_{mag} = \frac{\min(\max(\|\mathbf{x}_i\|, l_a), u_a) - l_a}{u_a - l_a},
\label{eq:mag_score}
\end{equation}
where $l_a$ and $u_a$ are the lower and upper bounds of the feature norm, respectively. The operation $\min(\max(\|\mathbf{x}_i\|, l_a), u_a)$ ensures that the feature norm is projected into the interval $[l_a, u_a]$, preventing extreme values from dominating the quality estimation. This linear normalization maps the bounded feature norm to a quality score. However, this does not cover identity-agnostic noise, which can degrade the discriminativeness of the representations.

\subsubsection{Semantic Quality Learning}
As shown in \cite{doan2025qcface}, embedding magnitude, such as \cite{meng2021magface,kim2022adaface}, can be significantly inflated by identity-agnostic artifacts such as motion blur, occlusion, or illumination variations, leading the model to assign high quality scores to unidentifiable samples.
To overcome this issue of identity-agnostic noise, we introduce a semantic quality learning approach based on squeeze-and-excitation semantic attention.
Given a face representation $\mathbf{x}$, a squeeze-and-excitation gating mechanism \cite{hu2018squeeze} is utilized to create an attended representation
\begin{equation}
\mathbf{x}_{att} = \mathbf{x} \odot \sigma(\mathbf{W}_2 \cdot \delta(\mathbf{W}_1 \mathbf{x})),
\label{eq:se_attention}
\end{equation}
with $\odot$ denotes element-wise multiplication, $\sigma(\cdot)$ is the sigmoid activation, $\delta(\cdot)$ is the GELU activation~\cite{hendrycks2016gaussian}, and $\mathbf{W}_1 \in \mathbb{R}^{\frac{d}{r} \times d}$, $\mathbf{W}_2 \in \mathbb{R}^{d \times \frac{d}{r}}$ are learnable parameters. The reduction ratio $r$ is a hyperparameter that controls the bottleneck dimension.
The attended representation $\mathbf{x}_{att}$ is then passed through a small multi-layer perceptron to produce a semantic quality score $q_{sem} \in [0,1]$
\begin{equation}
q_{sem} = \sigma\left(\mathbf{W}_4 \cdot \text{BN}\left(\delta(\mathbf{W}_3 \mathbf{x}_{att})\right)\right),
\label{eq:semantic_score}
\end{equation}
where $\mathbf{W}_3 \in \mathbb{R}^{d \times d}$, $\mathbf{W}_4 \in \mathbb{R}^{1 \times d}$, and BN denotes batch normalization. This score represents the estimated ``identity-utility'' of the feature, ensuring that only facial structures genuinely contributing to recognizability are prioritized during margin adaptation.

\subsubsection{Magnitude and Semantic Quality-based Margin Learning}
Lastly, an unified quality score $q$ 
\begin{equation}
q = \alpha \cdot q_{sem}  + (1 - \alpha) \cdot q_{mag},
\label{eq:fusion}
\end{equation}
is obtained through a simple combination of the semantic and magnitude-based quality scores.
Here, $\alpha \in [0,1]$ is a trade-off parameter balancing the physical reliability of the embedding norm with semantic consistency. 
As we will show later, this quality score $q$ is highly effective for FIQA within the DQM-Face framework. 
The fused quality score $q$ is then used to adaptively scale the target-class margin $m_1$:
\begin{equation}
m_1(q) = (u_m - l_m) \cdot q + l_m,
\label{eq:adaptive_margin}
\end{equation}
where $l_m$ and $u_m$ are the lower and upper bounds of the angular margin. This linear mapping ensures that higher-quality samples receive larger margins, pulling them closer to their class centers, while lower-quality samples are subject to weaker constraints, preventing overfitting on ambiguous features.

\subsection{Explicit Repulsion Margin Learning}
\label{sec:RepulsionMargin}
Most margin-based losses focus on intra-class compactness by enforcing a margin between samples and their ground-truth centers. However, this one-sided optimization can increase the risks of increase false matches as noted in \cite{liu2022sphereface,song2024coreface}.
Therefore, we explicitly enforce a repulsive margin $m_2$ to maximize discriminability.
In Eq \ref{eq:DQM}, the logit is defined as
\begin{equation}
z_j = s \cdot \cos(\theta_j - m_2),
\label{eq:repulsive_logit}
\end{equation}
with the repulsion margin $m_2 > 0$. This formulation creates an asymmetric decision boundary by actively pushing features away from incorrect class centers. Geometrically, it enforces a stricter angular ``clearance zone'' around each class, ensuring that samples not only cluster tightly around their own centers but also maintain distance from all other centers.

To improve optimization stability, we adopt a curriculum-based schedule for the inter-class repulsion margin $m_2$. Specifically, $m_2$ is progressively increased in three stages throughout training following:
\begin{equation}
m_2^{(t)} = 
\begin{cases} 
0.05, & 1 \leq t \leq 10 \\
0.10, & 11 \leq t \leq 18 \\
0.15, & 19 \leq t \leq 25
\end{cases}
\end{equation}
where $t$ is the current epoch. This staged strategy allows stable cluster formation during early epochs ($m_2 = 0.05$), followed by moderate inter-class separation in the intermediate stage ($m_2 = 0.10$), before enforcing a stronger repulsive margin ($m_2 = 0.15$) in the final stage. This progressive schedule prevents gradient instability while maximizing global separability in the embedding space.

\section{Experiments}

\subsection{Datasets, Training, and Evaluation}
Model training is conducted on the refined MS1MV2 dataset~\cite{deng2019arcface}, which contains approximately 5.8 million face images spanning 85,742 unique identities. Following standard protocols, all identities overlapping with evaluation benchmarks are removed to ensure unbiased open-set evaluation. Training is performed using Stochastic Gradient Descent (SGD) with a momentum of 0.9 and a weight decay of $5\times10^{-4}$. The initial learning rate is set to 0.1 and reduced by a factor of 10 at epochs 10, 18, and 22, with training terminating at epoch 25. The feature scale is fixed at $s = 64$, following standard practice in margin-based losses~\cite{deng2019arcface,wang2018cosface}. All models are trained with a batch size of 512.

For evaluating face recognition, we conduct experiments on widely adopted face verification benchmarks that cover diverse variations in pose, age, and image quality: LFW~\cite{huang2007labeled}, CFP-FP~\cite{sengupta2016frontal}, AgeDB-30~\cite{moschoglou2017agedb}, and CPLFW~\cite{zheng2018cplfw}. These benchmarks provide predefined face pairs that enable model comparison using the accuracy metric.
For more challenging large-scale evaluations, we further experiment with the IJB-B~\cite{whitelam2017iarpa} and IJB-C~\cite{maze2018iarpa} benchmarks, which include both still images and video frames with significant variations in resolution, occlusion, motion blur, and illumination. On these datasets, we follow the official 1:1 verification protocol and report the True Accept Rate (TAR) at different False Accept Rate (FAR) operating points, ranging from $10^{-3}$ to $10^{-5}$. Note that these are system-level metrics that account for preprocessing errors (e.g., detection and alignment), consistent with standard face recognition evaluation practices~\cite{iso19795}.

To evaluate Face Image Quality Assessment (FIQA) performance, experiments are conducted on four widely used face verification benchmarks: LFW~\cite{huang2007labeled}, Adience~\cite{6906255}, CPLFW~\cite{zheng2018cplfw}, and XQLFW~\cite{knoche2021xqlfw}. These datasets represent diverse challenges, including age variations (Adience), pose variations (CPLFW), severe image quality degradations (XQLFW), and an unconstrained in-the-wild setting (LFW). Performance is compared against nine state-of-the-art quality estimation methods, including eDifFIQA~\cite{10468647}, DifFIQA~\cite{babnik2023diffiqa}, CR-FIQA~\cite{boutros2023cr}, CLIB-FIQA~\cite{Ou_2024_CVPR}, GraFIQs~\cite{kolf2024grafiqs}, FROQ~\cite{babnik2025froq}, MagFace~\cite{meng2021magface}, SER-FIQA~\cite{Terhorst2020SERFIQUE}, and FaceQAN~\cite{babnik2022faceqan}. Evaluation is conducted using the Error-vs-Discard (EvD) Characteristics~\cite{isoFIQA}, also known as Error-vs-Reject curves~\cite{DBLP:conf/icb/TerhorstKDKK20}. In this procedure, images with the lowest predicted quality scores are progressively removed, and recognition performance is evaluated on the remaining subset. The resulting curves quantify how effectively a quality assessment method identifies and filters low-quality samples to reduce recognition errors.
For the error rate metric, we follow the recommendations of the European Border Control Agency (Frontex)~\cite{frontex2015abc}, computing the False Non-Match Rate (FNMR) at a fixed False Match Rate (FMR) of $10^{-3}$. To enable concise and comparable evaluation across methods, we additionally compute the partial Area Under the Curve (pAUC) of the EvD up to a 30\% rejection rate~\cite{DBLP:journals/tbbis/SchlettRTB24}. The pAUC represents the area under the EvD between 0\% and 30\% rejection; lower pAUC values indicate better FIQA performance.
All evaluations are performed using face embeddings extracted from the proposed DQM-Face model, since the predicted quality is model-specific and this ensures consistent comparisons across different FIQA methods.

\subsection{Architecture and Preprocessing}

For the backbone network, a iResNet-100 architecture~\cite{duta2020improved} is used, which extracts 512-dimensional face embeddings. The semantic-attention branch is implemented as a lightweight module inserted after the global average pooling layer, consisting of a Squeeze-and-Excitation (SE) block with a reduction ratio of 16, followed by a two-layer MLP with hidden dimension 512.
For preprocessing, we follow the InsightFace framework~\cite{insightface}.

\subsection{Magnitude Hyperparameter Settings}
The main hyperparameters used in DQM-Face follow standard practices from margin-based face recognition and refinements by \cite{meng2021magface}. The quality-dependent target margin is linearly interpolated between a lower bound of 0.35 and an upper bound of 0.8. The feature magnitude bounds are set to 10 and 110, with the magnitude regularization weight fixed at 35. Otherwise, the quality fusion weight is set to 0.5, balancing the contributions of magnitude- and semantic-based quality components. The maximum inter-class repulsive margin is fixed at 0.15 and follows a curriculum scheduling strategy during training. The choice of these is justified in the ablation studies.

\section{Results}

\subsection{Recognition Performance on Single-Image Benchmarks}

To assess the effectiveness of the proposed DQM-Face method, Table~\ref{tab:ms1m_arcface_r100_results} presents its verification performance in comparison with state-of-the-art face recognition approaches on standard image-to-image benchmarks. These evaluations cover challenging scenarios such as cross-pose and cross-age matching. All models were trained on the MS1MV2 dataset \cite{guo2016ms} using an iResNet-100 backbone \cite{duta2020improved}. The results indicate that DQM-Face consistently achieves superior performance, ranking first or second across all evaluated benchmarks, demonstrating the effectiveness of the quality-guided dual margin. We attribute the consistent performance observed across these challenging data pairs to the integration of quality estimation into the learning process, which facilitates context-dependent representation learning.

However, as shown in \cite{fallahi2025reliability}, these benchmarks are relatively small in size, and the resulting performance differences are so minor that they cannot be considered statistically significant. Therefore, we conduct additional experiments on the large-scale IJB-B and IJB-C benchmarks to provide a more robust evaluation.

\begin{table*}[t]
\centering
\caption{\textbf{Recognition Performance on Single-Image Benchmarks -} The highest verification accuracy (\%) is highlighted in bold, and the second-best result is underlined. The proposed DQM-Face solution performs well across the different benchmarks.}
\label{tab:ms1m_arcface_r100_results}
%\resizebox{\textwidth}{!}{%
\begin{tabular}{lcccccc}
\toprule
\textbf{Method} & \textbf{LFW} & \textbf{AgeDB-30} & \textbf{CFP-FP} & \textbf{CPLFW} & \textbf{Avg.} \\
\midrule
SphereFace~\cite{liu2017sphereface} & 99.67 & 97.05 & 96.84 & 91.27 & 96.21 \\
CosFace~\cite{wang2018cosface} & 99.81 & 98.12 & 98.12 & 92.48 & 97.13 \\
ArcFace~\cite{deng2019arcface} & \underline{99.83} & 98.15 & 98.40 & 92.72 & 97.28 \\
GroupFace~\cite{kim2020groupface} & \textbf{99.85} & 98.28 & 98.63 & 93.17 & 97.48 \\
CurricularFace~\cite{huang2020curricularface} & 99.80 & 98.32 & 98.37 & 93.13 & 97.41 \\
MagFace~\cite{meng2021magface} & \underline{99.83} & 98.17 & 98.46 & 92.87 & 97.33 \\
AdaFace~\cite{kim2022adaface} & 99.82 & 98.05 & 98.49 & \textbf{93.53} & 97.47 \\
ElasticFace-Arc+~\cite{boutros2022elasticface} & 99.82 & 98.35 & 98.67 & 93.28 & \underline{97.53} \\
ElasticFace-Cos+~\cite{boutros2022elasticface} & 99.80 & 98.28 & 98.73 & 93.23 & 97.51 \\

QMagFace~\cite{terhorst2023qmagface} & \underline{99.83} & \textbf{98.50} & \underline{98.74} & - & - \\

CoReFace~\cite{song2024coreface} & - & \underline{98.37} & 98.60 & 93.27 & - \\
QCFace~\cite{doan2025qcface} & 99.80 & 98.18 & 98.46 & 92.83 & 97.31 \\
\midrule

DQM-Face $\alpha_{0.4}$ (Ours)   & \underline{99.83} & 98.32 & 98.61 & 93.22 & 97.50 \\

DQM-Face $\alpha_{0.5}$ (Ours)   & \underline{99.83} & 98.35 & \textbf{98.83} & \underline{93.30} &\textbf{97.58}\\
\bottomrule
\end{tabular}
\end{table*}

\subsection{Recognition Performance on Video-based Benchmarks}

To evaluate the performance of the proposed DQM-Face method on large-scale datasets, Table~\ref{tab:ijb_results} reports the TAR at several FARs on the widely used IJB-B and IJB-C benchmarks. The results, compared with those of state-of-the-art approaches, demonstrate that DQM-Face consistently achieves the best or second-best performance across all evaluated FARs. We attribute the strong performance in higher FAR regions (\(10^{-3}\)) to the quality-guided learning, as face image quality factors have a pronounced impact in this regime~\cite{terhorst2023qmagface}. Similarly, the high performance in lower FAR regions (\(10^{-5}\)) may be attributed to the dual-margin learning, as enhanced identity separability is expected to be beneficial when stricter discrimination is required.

\begin{table*}[t]
\centering
\caption{\textbf{Recognition Performance on Video-based Benchmarks -} Results are reported as True Acceptance Rate (TAR, \%) at various False Acceptance Rate (FAR) thresholds. The best and second-best results are highlighted in bold and underlined, respectively. The proposed DQM-Face solution achieves state-of-the-art performance across different FARs on both, IJB-B and IJB-C.
}
\label{tab:ijb_results}
%\resizebox{\textwidth}{!}{%
\begin{tabular}{lcccccc}
\toprule
& \multicolumn{3}{c}{\textbf{IJB-B}} & \multicolumn{3}{c}{\textbf{IJB-C}} \\
\cmidrule(lr){2-4} \cmidrule(lr){5-7}
\textbf{Method} & $10^{-3}$ & $10^{-4}$ & $10^{-5}$ & $10^{-3}$ & $10^{-4}$ & $10^{-5}$ \\
\midrule
CosFace~\cite{wang2018cosface} & - & 94.01 & 89.25 & - & 95.56 & 92.68 \\
ArcFace~\cite{deng2019arcface} & - & 94.09 & 88.50 & - & 95.74 & 92.69 \\
PFE~\cite{shi2019probabilistic} & - & - & - & 95.49 & 93.25 & 89.64 \\
GroupFace~\cite{kim2020groupface} & - & 94.93 & - & - & 96.26 & - \\
CurricularFace~\cite{huang2020curricularface} & - & 94.80 & - & - & 96.10 & - \\
EQFace~\cite{liu2021eqface} & 96.48 & 94.51 & 89.62 & 97.39 & 95.84 & 93.01 \\
%EQFace* (QWFA)~\cite{liu2021eqface} & 96.31 & 94.88 & \textbf{91.87} & 97.45 & 96.38 & \textbf{94.93} \\
MagFace~\cite{meng2021magface} & 96.19 & 94.50 & 90.36 & 97.24 & 95.97 & 94.08 \\
SCF-ArcFace~\cite{li2021spherical} & - & 94.74 &90.68 & - & 96.09 & 94.04 \\
%ElasticFace-Arc+~\cite{boutros2022elasticface} & - & 95.22 & - & - & 96.49 & - \\
ElasticFace-Cos+~\cite{boutros2022elasticface} & - & \underline{95.30} & - & - & \underline{96.57} & - \\
SphereFace-R v2~\cite{liu2022sphereface} & - & 94.51 & 86.55 & - & 95.96 & 93.01 \\
DDC~\cite{uzun2022deep} & - & 94.70 & - & - & 96.10 & - \\
QMagFace~\cite{terhorst2023qmagface} & 96.48 & 94.70 & 90.29 & 97.62 & 96.19 & 94.27 \\
Face-TB~\cite{zhu2023joint} & - & 94.37 & - & - & 95.72 & - \\

CoReFace~\cite{song2024coreface} & - & 95.09 & \underline{91.33} & - & 96.43 & 94.73 \\
QCFace~\cite{doan2025qcface} & - & 94.30 & 89.39 & - & 95.84 & 93.82 \\
\midrule
DQM-Face $\alpha_{0.4}$ (Ours)  & \textbf{96.71} & \textbf{95.38} & \textbf{91.35} & \textbf{97.65} & \textbf{96.58} & \textbf{94.83} \\

DQM-Face $\alpha_{0.5}$ (Ours)  & \underline{96.65} & 95.20 & 90.68 & \underline{97.64} & 96.53 & \underline{94.78} \\

%DQM-Face  (Ours)  & \textbf{96.65} & \underline{95.20} & \underline{90.68} & \textbf{97.64} & \underline{96.53} & \textbf{94.78} \\

\bottomrule
\end{tabular}
\end{table*}

\subsection{Face Image Quality Assessment Performance}

\begin{figure*}[tb]
    \centering
    
    \begin{subfigure}[b]{0.46\textwidth}
        \vspace{0.2em}
        \centering
        \includegraphics[width=\textwidth]{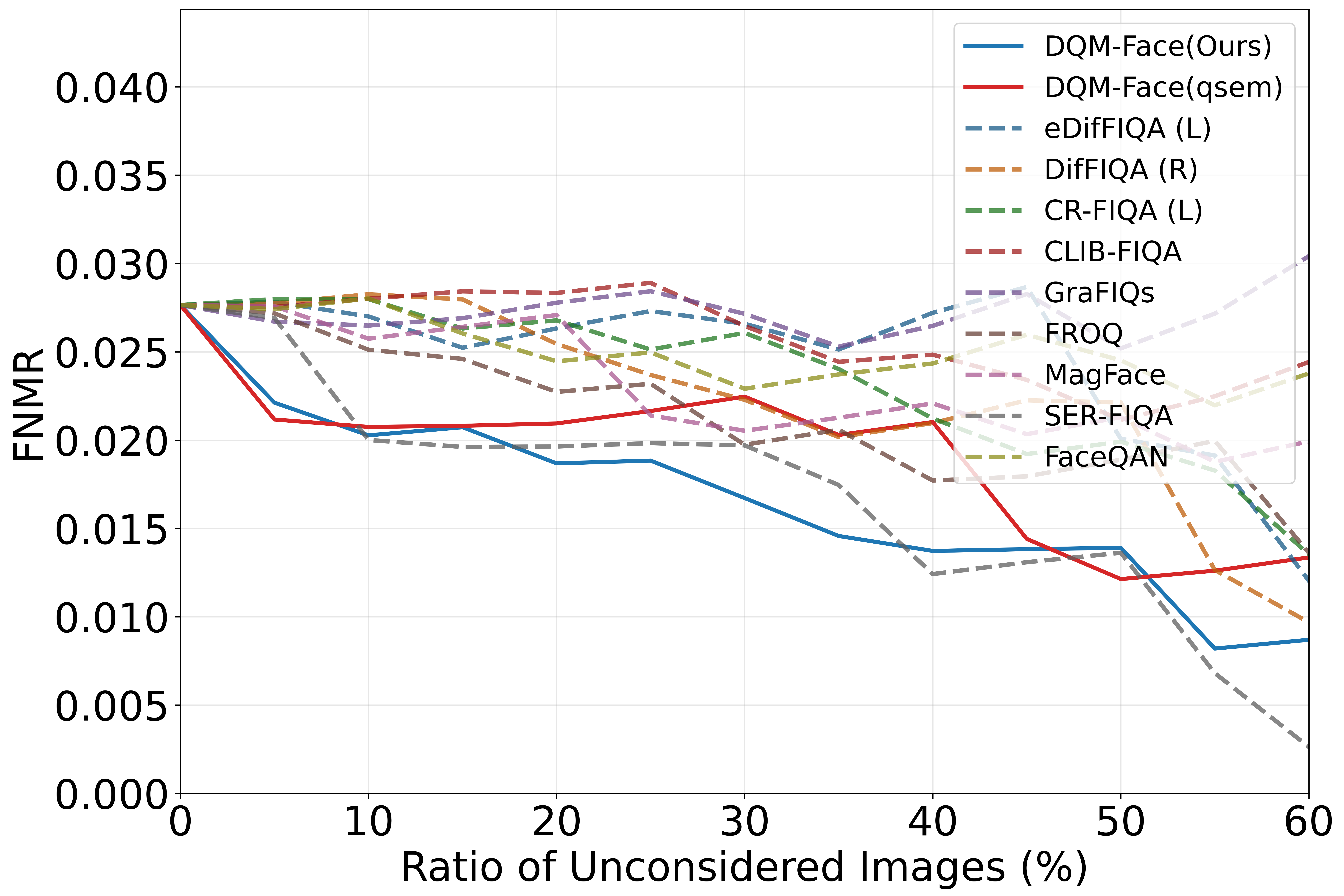}
        \caption{LFW (unconstrained)}
        \label{fig:fiq_lfw}
    \end{subfigure}
    \hfill
    \begin{subfigure}[b]{0.46\textwidth}
        \vspace{0.2em}
        \centering
        \includegraphics[width=\textwidth]{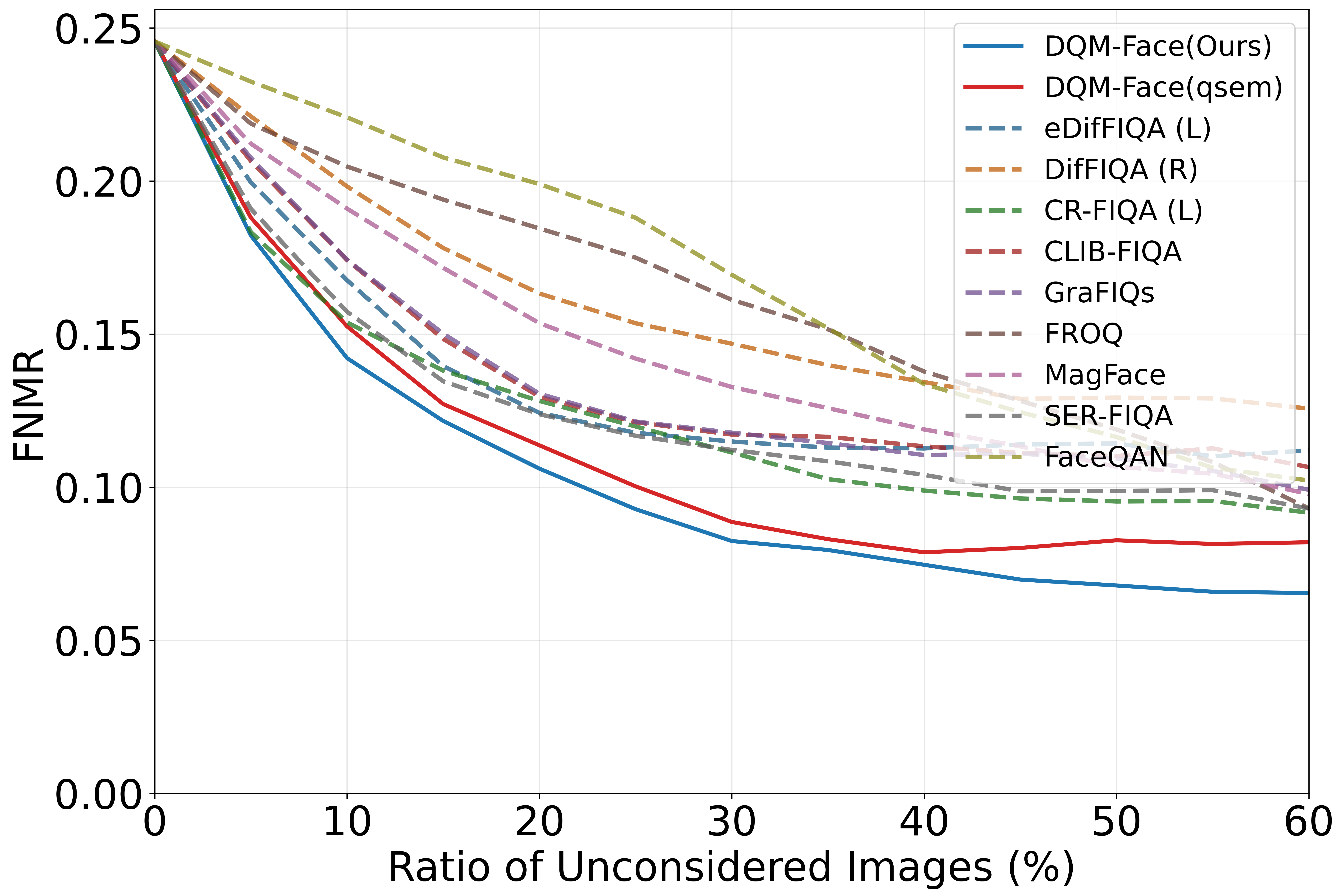}
        \caption{Adience (age variations)}
        \label{fig:fiq_adience}
    \end{subfigure}

     \begin{subfigure}[b]{0.46\textwidth}
        \vspace{0.2em}
        \centering
        \includegraphics[width=\textwidth]{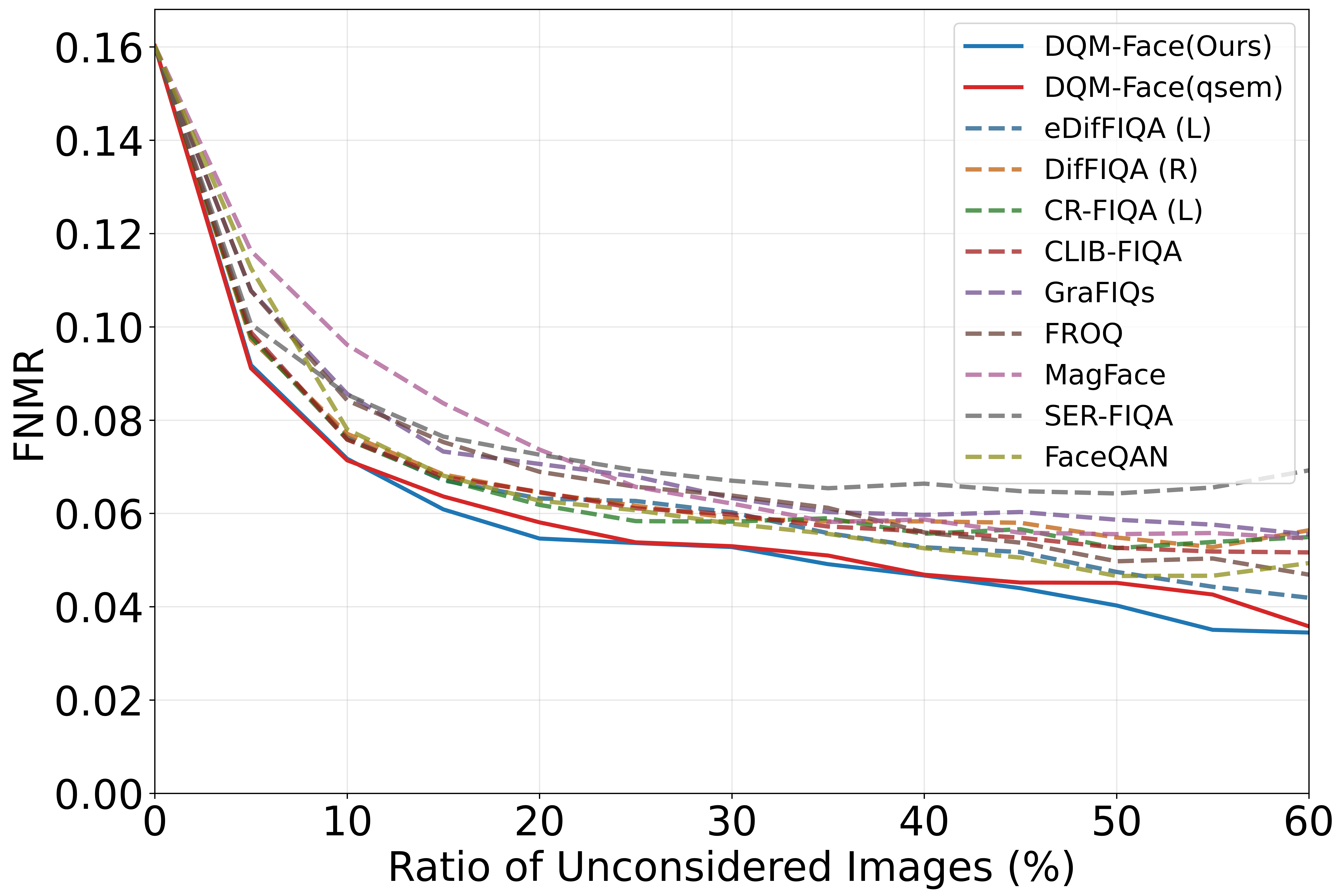}
        \caption{CP-LFW (pose variations)}
        \label{fig:fiq_cplfw}
    \end{subfigure}
    \hfill
    \begin{subfigure}[b]{0.46\textwidth}
        \vspace{0.2em}
        \centering
        \includegraphics[width=\textwidth]{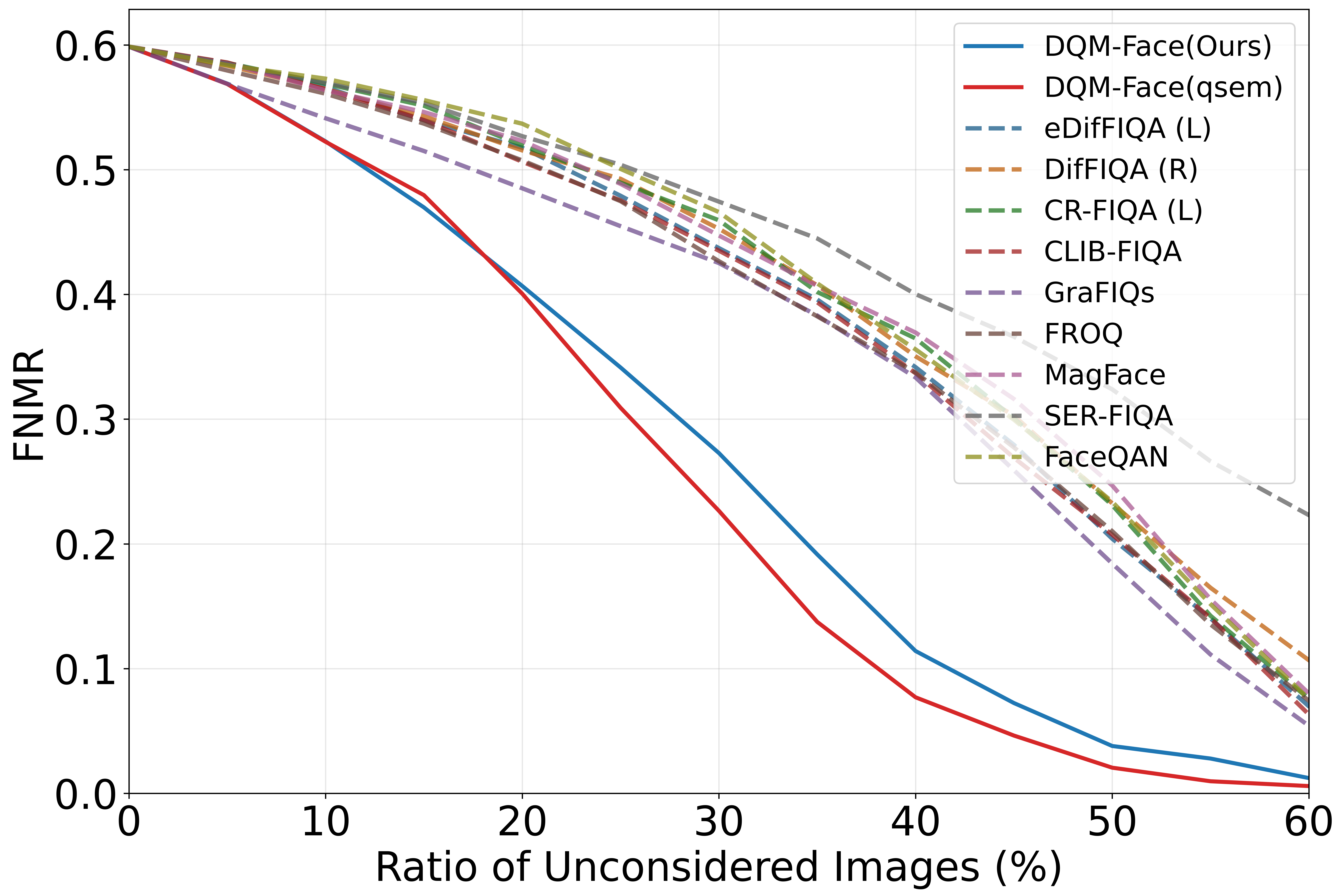}
        \caption{XQLFW (image quality variations)}
        \label{fig:fiq_xqlfw}
    \end{subfigure}
    
    \caption{\textbf{Face Image Quality Assessment Performance -} The FIQA performance is evaluated using Error-vs-Discard curves on four popular FIQA datasets with specific challenges. The proposed method is compared against nine state-of-the-art approaches on the DQM-Face embeddings, demonstrating that its predicted quality scores effectively reflect the underlying sample utility for the DQM-Face recognition model.}
    \label{fig:comparison_roc}
\end{figure*}

To measure the effectiveness of the proposed quality estimate utilized for the margin learning, the FIQA performance is investigated using EvD characteristics and pAUC values on four standard datasets, each posing unique challenges: LFW (unconstrained), Adience (age variations), CPLFW (pose variations), and XQLFW (image quality variations). 
Figure~\ref{fig:comparison_roc} illustrates the EvD characteristics, while Table~\ref{tab:fiq_results} summarizes the pAUC values computed at a 30\% discard rate. 
For a comprehensive evaluation, the proposed method is compared against nine state-of-the-art FIQA approaches.

The DQM-Face model demonstrates consistently strong performance, ranking among the top-performing methods across all datasets. 
It achieves the best results on LFW, Adience, and CPLFW, and remains highly competitive on XQLFW. 
To further analyze the role of semantic quality, we introduce the DQM-Face (qsem) variant, trained with $\alpha = 1$ to capture only the semantic quality component $q_{sem}$.
This variant achieves the highest performance on XQLFW and maintains competitive results on the remaining datasets, indicating that the semantic quality contributes to robustness under severe quality variations.
Although the FIQA performance is strong, the quality learning mechanism is deeply integrated into the face recognition framework to capture subtle, identity-preserving variations. Consequently, the learned quality is inherently model-specific, relying on architecture-dependent cues rather than aiming for universal generalization, yet it still achieves strong overall performance.

\begin{table*}[tb]
\centering
\caption{
\textbf{Face Image Quality Assessment Evaluation -} is evaluated using pAUC, where lower values indicate better performance. 
The best result is highlighted in bold, and the second-best is underlined. 
Evaluations are conducted on widely used FIQA datasets.
%: LFW (unconstrained), Adience (age variations), CPLFW (pose variations), and XQLFW (image quality variations). 
The proposed approach is compared against nine state-of-the-art methods using the DQM-Face embeddings, demonstrating that its predicted quality scores effectively capture its underlying sample utility.}
\label{tab:fiq_results}
\begin{tabular}{lcccc}
\toprule
\textbf{Method} & \textbf{LFW} & \textbf{Adience} & \textbf{CPLFW} & \textbf{XQLFW} \\
\midrule
DQM-Face (Ours) & \textbf{0.0061} & \textbf{0.0404} & \textbf{0.0220} & \underline{0.1373} \\
DQM-Face (qsem) & \underline{0.0065} & \underline{0.0424} & \underline{0.0222} & \textbf{0.1346} \\
eDifFIQA (L) \cite{10468647} & 0.0080 & 0.0465 & 0.0239 & 0.1603 \\
DifFIQA (R) \cite{babnik2023diffiqa} & 0.0079 & 0.0555 & 0.0239 & 0.1612 \\
CR-FIQA (L) \cite{boutros2023cr} & 0.0081 & 0.0451 & 0.0235 & 0.1622 \\
CLIB-FIQA \cite{Ou_2024_CVPR} & 0.0084 & 0.0481 & 0.0239 & 0.1594 \\
GraFIQs \cite{kolf2024grafiqs} & 0.0082 & 0.0483 & 0.0258 & 0.1538 \\
FROQ \cite{babnik2025froq} & 0.0073 & 0.0590 & 0.0257 & 0.1586 \\
MagFace \cite{meng2021magface} & 0.0076 & 0.0530 & 0.0273 & 0.1615 \\
SER-FIQA \cite{Terhorst2020SERFIQUE} & 0.0065 & 0.0451 & 0.0259 & 0.1638 \\
FaceQAN \cite{babnik2022faceqan} & 0.0078 & 0.0628 & 0.0246 & 0.1641 \\
\bottomrule
\end{tabular}
\end{table*}

\subsection{Visual Quality Analysis}
\begin{figure}[h]
    \centering
    \includegraphics[width=0.88\linewidth]{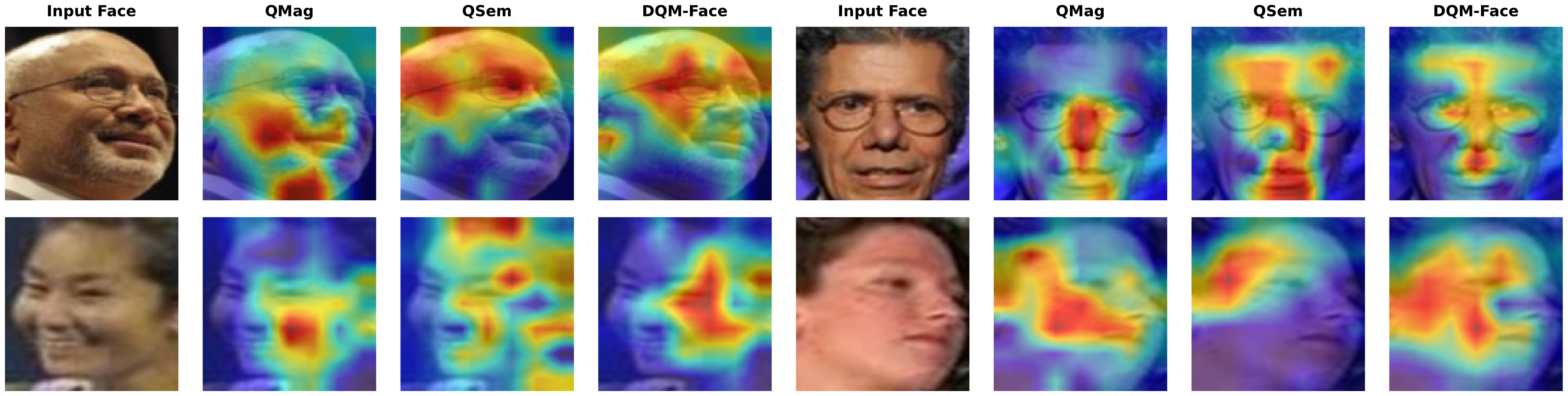}
    \caption{\textbf{Component-wise Visual Attribution Analysis -} Comparison of Grad-CAM heatmaps for the three model variants:  $q_{mag}$ (magnitude-only, $\alpha = 0$), $q_{sem}$ (semantic-only, $\alpha = 1$), and DQM-Face (fused quality, $\alpha = 0.5$). The heatmaps highlight the facial regions that contribute most to identity recognition. }
    \label{fig:grad_cam}
\end{figure}

To provide qualitative insight into the different quality branches, we employ Grad-CAM~\cite{selvaraju2017grad} to visualize the spatial regions that contribute most to the recognition decision. Figure~\ref{fig:grad_cam} compares Grad-CAM heatmaps for the three model variants: $q_{mag}$ (magnitude-only, $\alpha=0$), $q_{sem}$ (semantic-only, $\alpha=1$), and the proposed DQM-Face (fused quality, $\alpha=0.5$). The heatmaps are overlaid on the input face images to illustrate the regions emphasized by each model. The $q_{mag}$ variant exhibits attention that is occasionally dispersed toward non-discriminative regions, such as background textures or facial boundaries, particularly when feature magnitudes are affected by blur, occlusion, or low image quality. This observation supports the hypothesis that feature magnitude alone cannot reliably represent identity quality. In contrast, the $q_{sem}$ variant attends to identity-discriminative facial regions more consistently, including the eyes, nose, and mouth.. This behavior demonstrates that the semantic quality branch successfully learns identity-aware quality representations that are more closely aligned with the face recognition objective. The proposed DQM-Face model ($\alpha=0.5$) combines the strengths of both quality cues. Its attention maps exhibit well-localized and stable activation over the most informative facial regions while remaining robust to challenging imaging conditions. Compared with either individual branch, the fused model produces more focused and semantically meaningful attention, indicating that magnitude-based and semantic quality provide complementary information.

\subsection{Complexity Analysis}
The complexity of DQM-Face is comparable to similar FR solutions, such as ArcFace.
The SE branch adds $+0.099$M parameters ($+0.15\%$) and less than $0.0001$ GFLOPs on the backbone (of 12.149 GFLOPs, 65.156M parameters). This overhead is negligible compared to ArcFace~\cite{deng2019arcface} ($<0.2\%$). The inference time is dominated by the used backbone, as $m_2$ adds no cost and quality estimation is based directly on the extracted features.

\subsection{Ablation Studies}

To validate the choice of hyperparameters, we conducted an ablation study investigating different configurations. Since the face recognition performance in our approach depends on the quality margin $m_1(q(\alpha))$, which is influenced by the hyperparameter and the quality fusion weight $\alpha$, we analyzed how varying $\alpha$ affects recognition accuracy. Table~\ref{tab:ablation_alpha_updated} summarizes the results for $\alpha$ values across the full range $[0, 1]$.
%On single-image benchmarks, 
The model achieves its highest accuracy with $\alpha = 0.5$, while performance gradually decreases toward the extreme values. 

Based on these observations, we adopt $\alpha = 0.5$ as the default setting for the proposed method. Overall, these findings support our hypothesis that fusing complementary quality cues yields a more robust assessment of sample utility than relying on either component in isolation.

\begin{table*}[tb]
\centering
\footnotesize
\setlength{\tabcolsep}{2pt}
\renewcommand{\arraystretch}{0.8}
\caption{\textbf{Ablation study on quality margin} – The effect of the quality margin $m_1$ through the variations of the fusion weight $\alpha$ is studied. $\alpha = 0$ corresponds to magnitude-based quality only ($q_{mag}$), while $\alpha = 1$ corresponds to semantic quality only ($q_{sem}$).}
\label{tab:ablation_alpha_updated}
\begin{tabular}{cccccccccccc}
\toprule
& \multicolumn{5}{c}{Verification Accuracy (\%)} & \multicolumn{3}{c}{IJB-B (TAR@FAR)} &
\multicolumn{3}{c}{IJB-C (TAR@FAR)} \\
\cmidrule(lr){2-6}
\cmidrule(lr){7-9}
\cmidrule(lr){10-12}

$\alpha$ &LFW &CFP-FP &AgeDB &CPLFW &Avg. &$10^{-3}$ &
$10^{-4}$ &$10^{-5}$ &$10^{-3}$ &$10^{-4}$ &$10^{-5}$ \\
\midrule
0.0 & 99.81 & 98.41 & 98.31 & 93.00 & 97.38 & 96.49 & 95.01 & 90.58 & 97.61 & 96.40& 94.45 \\

0.1 & 99.83 & 98.24 & 98.20 & 93.08 & 97.34 & 96.56 & 94.99 & \underline{91.19} & 97.54 & 96.31 & 94.64 \\

0.2 & 99.83 & 98.50 & 98.27 & 93.02 & 97.41 & \textbf{96.77} & \underline{95.25} & 90.97 & 97.63 & 96.43 & 94.49 \\

0.3 & 99.83 & 98.44 & 98.20 & 93.05 & 97.38 & 96.48 & 94.80 & 89.70 & 97.43 & 96.20 & 94.34 \\

0.4 & 99.83 & 98.61 & \underline{98.32} & \underline{93.22} & \underline{97.50} & \underline{96.71} & \textbf{95.38} & \textbf{91.35} & \textbf{97.65} & \textbf{96.58} & \textbf{94.83} \\

0.5 & \textbf{99.83} & \textbf{98.83} & \textbf{98.35} & \textbf{93.30} & \textbf{97.58} & 96.65 & 95.20 & 90.68 & \underline{97.64} & \underline{96.53} & \underline{94.78} \\

0.6 & 99.82 & 98.67 & 98.25 & 93.02 & 97.44 & 96.57 & 94.82 & 89.95 & 97.55 & 96.21 & 94.14 \\

0.7 & 99.80 & 98.57 & 98.25 & 93.20 & 97.46 & 96.67 & 94.97 & 89.28 & 97.56 & 96.32 & 94.33 \\

0.8 & 99.82 & 98.34 & 98.20 & 93.10 & 97.37 & 96.54 & 94.81 & 90.46 & 97.55 & 96.23 & 94.22 \\

0.9 & 99.83 & 98.36 & 98.28 & 93.07 & 97.39 & 96.48 & 94.95 & 89.20 & 97.52 & 96.38 & 94.24 \\

1.0 & 99.82 & \underline{98.69} & 98.22 & 93.17 & 97.48 & 96.67 & 95.22 & 90.55 & 97.62 & 96.51 & 94.67 \\

\bottomrule
\end{tabular}
\end{table*}

Table~\ref{tab:ablation_m2} analyzes the effect of the repulsion margin hyperparameter $m_2$. 
For training DQM-Face, a curriculum-based scheduling strategy for $m_2$ was introduced in Section~\ref{sec:RepulsionMargin}. 
To evaluate its impact, Table~\ref{tab:ablation_m2} reports the performance obtained with different fixed margin values, ranging from $m_2 = 0$ (no inter-class repulsion) to $m_2 = 0.2$ (strong fixed repulsion).
We additionally compare against linear $ m_2(t) = 0.15 \cdot t/T $ and cosine $ m_2(t) = 0.15 \cdot (1-\cos(\pi t/T))/2 $ schedules. While both achieve competitive performance, they underperform the staged schedule. The staged design maintains constant $m_2$ during each phase, providing stable gradients and better alignment with curriculum learning.
The results indicate that the proposed curriculum-based schedule yields the best performance, as it enables the model to first form compact identity clusters and then progressively enforce stronger inter-class separation. 
This gradual adaptation stabilizes the gradient, and thus its training process, and promotes higher global separability among the face representations.

\begin{table*}[tbh]
\centering
\footnotesize
\setlength{\tabcolsep}{2pt}
\renewcommand{\arraystretch}{0.8}
\caption{\textbf{Ablation study on the repulsion margin} - The effect of the repulsion $m_2$, and specifically its scheduled learning procedure, is studied.}
\label{tab:ablation_m2}
\begin{tabular}{lccccccccccc}
\toprule
& \multicolumn{5}{c}{Verification Accuracy (\%)} & \multicolumn{3}{c}{IJB-B (TAR@FAR)} &
\multicolumn{3}{c}{IJB-C (TAR@FAR)} \\
\cmidrule(lr){2-6}
\cmidrule(lr){7-9}
\cmidrule(lr){10-12}
$m_2$ &LFW &CFP-FP &AgeDB &CPLFW &Avg. &$10^{-3}$ &
$10^{-4}$ &$10^{-5}$ &$10^{-3}$ &$10^{-4}$ &$10^{-5}$ \\
\midrule
$0$ 
& 99.83 & 98.43 & 98.13 & 92.92 & 97.32 &96.63&94.87&87.21&97.58&96.19 &93.75 \\
$0.05$ & 99.81 & 98.64 & 98.18 & \underline{93.26} & \underline{97.47} & 96.54&94.94 &89.85  &97.57 &96.32  &94.31 \\
$0.10$ & 99.80 & \underline{98.71} & 98.13 & 93.23 & 97.46 & \textbf{96.78} & 95.05&90.04 &97.64  & 96.43&\underline{94.69}  \\
$0.15$ & 99.83 & 98.57 & 98.22 & 93.20 & 97.46 &96.67 & 95.08&88.80&\textbf{97.66} &\underline{96.52 }&94.53 \\
$0.20$ &99.33 & 85.96 & 95.05 & 85.63 & 91.49 & 91.60 & 86.91 & 76.69 & 93.38 & 89.28 & 83.37 \\
\midrule
Linear& 99.83 & 98.35 & \underline{98.28} & 93.06 & 97.38 &\underline{96.77} &\textbf{95.25} &\textbf{90.97} &97.63 &96.43&94.49 \\
 Cosine& 99.83 & 98.50&98.26 & 93.01 & 97.40& 96.48&94.95&89.20 &97.52 &96.38 &94.24 \\

\midrule
Staged  
& \textbf{99.83} & \textbf{98.83} & \textbf{98.35} & \textbf{93.30} & \textbf{97.58} & 96.65 & \underline{95.20} & \underline{90.68} & \underline{97.64} & \textbf{96.53} & \textbf{94.78} \\
\bottomrule

\end{tabular}
\end{table*}

\section{Conclusion}
In this work, we introduced DQM-Face, a Dual Quality Margin Learning framework designed to address the challenges of face recognition in unconstrained environments. By jointly modeling magnitude-based and semantic quality cues through a unified attraction–repulsion margin optimization, the proposed approach enables the learning of more discriminative and structured feature representations. The quality derived from margin optimization achieves high FIQA performance within the proposed framework, demonstrating its intrinsic alignment with the underlying recognition objective.
Comprehensive evaluations across diverse image-to-image and video benchmarks demonstrate that DQM-Face consistently achieves the best or second-best face recognition performance compared to state-of-the-art methods. These results confirm the effectiveness of integrating semantic quality modeling into margin-based learning.

\vspace{0.6cm}

\textbf{Acknowledgement.} This work was funded by the Deutsche Forschungsgemeinschaft (DFG, German Research Foundation) under Grant 544631027.

\clearpage

\bibliographystyle{splncs04}
\bibliography{main}
\end{document}